# IMAGE FILTERING USING ALL NEIGHBOR DIRECTIONAL WEIGHTED PIXELS: OPTIMIZATION USING PARTICLE SWARM OPTIMIZATION

## J. K. Mandal[1] and Somnath Mukhopadhyay[2]

[1]Department of Computer Engineering, University of Kalyani, Kalyani, West Bengal, India
jkm.cse@gmail.com
[1]Department of Computer Engineering, University of Kalyani, Kalyani, West Bengal, India
som.cse@live.com

## ABSTRACT

*In this paper a novel approach for de noising images corrupted by random valued impulses has been proposed. Noise suppression is done in two steps. The detection of noisy pixels is done using all neighbor directional weighted pixels (ANDWP) in the 5 x 5 window. The filtering scheme is based on minimum variance of the four directional pixels. In this approach, relatively recent category of stochastic global optimization technique i.e., particle swarm optimization (PSO) has also been used for searching the parameters of detection and filtering operators required for optimal performance. Results obtained shows better de noising and preservation of fine details for highly corrupted images.*

## 1. INTRODUCTION

Due to a number of non idealistic encountered in image sensors and communication channels digital images are often corrupted by impulses during image acquisition or transmission. In most of the image processing applications, the most important stage is to remove the impulses because the subsequent tasks such as segmentation, feature extraction, object recognition, etc. are affected by noises [1]. Various filtering methods have been suggested for the removal of impulses from the digital images. Most of these methods are based on median filtering techniques, which use the rank order information of the pixels in the filtering window. The *standard median filter* [1] removes the noisy pixels by replacing test pixel with the median value of the pixels in the window. This technique provides a standard noise removal performance but also removes thin lines and dots, distorts edges and blurs image fine textures even at low noise ratios. The *weighted median filter* [2], *center weighted median filter* [3] and *adaptive center weighted median filter* [4] are modified median filters. They give extra weight to some pixels of the filtering window and thus these filters achieve betterment to the standard median filter.

The standard and weighted median filters are incapable of making distinction between the noisy and noise less pixels of the noisy image. Hence these filters distort the noise free pixels of the image. For such problems, switching *median filter* [5] has been proposed in which an impulse detector has been introduced to classify the center pixel of the test window. If the center pixel is detected as noisy then that pixel is replaced by standard median value of the test window. Otherwise the window is not filtered. The performance of this method of filtering purely depends on the performance of impulse detection algorithm but this method of filtering noisy image performs considerably better to standard and weighted median filters. As a result, many impulse detection methods along with switching median filters have been proposed [4] - [8]. Among them, an iterative *pixel-wise modification of MAD* (median of the absolute deviations from the median) filter [8] is a robust estimator of the variance used to efficiently separate noisy pixels from the image details. The *tri-state median filter* [9] and *multistate median filter* [10] are improved switching median filters those are made using a weighted median filter and an appropriate number of center weighted median filters. These filters perform better than weighted and center median filters at the cost of increased computational complexity. The *progressive switching median filter* [11] is also a variant of switching median filter that recursively performs the impulse detection and removal in two different stages. This filter performs better than many other median filters but it has a very high computational cost due to its recursive nature. The *partition based median filter* [12] is an adaptive median filter has been introduced to tackle both impulse noise and Gaussian noise, which uses the LMS algorithm for optimization purpose. The *signal dependent rank ordered mean filter* [13] is a switching mean filter that uses rank order information for impulse detection and filter. This method is similar to the switching median filter except that the median operation is replaced with a rank ordered mean operation. This filter obtains better noise suppression quality than some state-of-the-art impulse noise removal techniques for both gray and color images. To deal with random valued impulse noises in the images, an advance median filter, *directional weighted median filter* [14] has been proposed. This scheme uses a new impulse detection method and which is based on the differences between the test pixel and its 16 neighborhood pixels aligned with four main directions in the 5 x 5 window. The filtering scheme used here is a variant of median filter. It iterates the detection and filtering algorithm a minimum of 8 to 10 times to give satisfactory results for the images having highly random valued noises. Another switching median filter developed by Sa, Dash and Majhi, the *second order difference based impulse detection filter* [15] takes all the neighborhood pixels in the 3 x 3 window to detect and filter the random valued impulse noises in the image. This method of removing impulses has a drawback that it does not work well for highly corrupted images but good for very low rate of impulses in images. *ANDWP* [22] filter has varied the user parameters in a particular range and searched them manually in the 3 dimensional space to optimize the operator. Although it is a difficult task to determine the best parameter set to optimize the results for the various images having different noise density. Hence in this paper we used a global optimization technique, PSO to determine and optimize the restoration results.

In addition to the median and mean based filters discussed, a dozens of soft computing tools based filters have also been proposed in this literature such as fuzzy filter [16], neuro fuzzy filter [17]., etc. These filters perform relatively better in terms of noise removal and details preservation compared to median and mean based filters. During noise suppression, a majority of the above mentioned filters have more or less drawbacks of removing thin lines and edges and thus blurring the fine textures in the images. Although these methods work fine for the images corrupted with impulses with up to 30% noise level in the images. But when more percentage of impulses presents in the images, these median and other filters don't able to perform satisfactory and they also can't remove some black patches on the reconstructed image.

In this paper the scheme for removal of random valued impulse noise has been proposed which uses all the neighborhood pixels for noise detection as well as for noise filtering in the 5 x 5 window. The method uses maximum possible information of the neighborhood in order to improve the quality of the reconstructed image. The filtering operator is based on minimum variance of the four directional pixels aligned in the 5 x 5 window. Three user parameters such as number of iterations (I), threshold value (T) and decreasing rate (R) of threshold value in each iteration are searched in a 3-Dimentional space to get global optimal solution using a stochastic search strategy, particle swarm optimization (PSO) technique. The performance of the proposed algorithm is experimented and compared with other methods under several noise densities and different bench mark images. Experimental results show that the proposed algorithm performs better noise suppressing quality and effective image fine details preservation.

Rest of the paper organization is as follows. Section 2 illustrates the impulse detection operator. Section 3 explains the filtering strategy. The proposed particle swarm optimization based technique is given in section 4.0. PSO based experiment results, comparisons and discussions are given in Section 5.0. Section 6.0 presents concluding remarks.

## 2. IMPULSE DETECTOR

### 2.1. Random Valued Impulse Noise

The images corrupted by impulsive noises with probability p can be described as follows:

$$X(k) = \begin{cases} n(k) \text{ with probability } p \\ \\ f(k) \text{ with probability } 1-p \end{cases}$$

Where n (k) denotes the image contaminated by impulse with noise ratio p, and f (k) means the pixels are noise free. There are two types of the impulsive noises: fixed- and random-valued impulses. In a gray-scale image, the fixed-valued impulse, known as salt and pepper noise, shows up as either 0 or 255 with equal probability (i.e. p/2), while the random-valued impulse is uniformly distributed over the range of[0, 255] at probability p.

### 2.2. Detection Rule

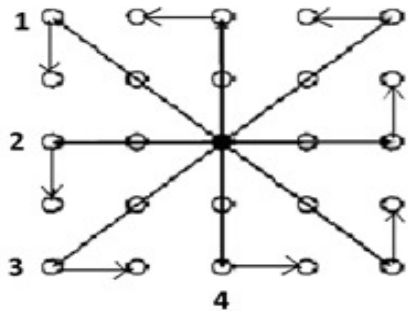

Fig. 1 Four Directional Weighted Pixels in the 5 x 5 window for impulse detection

In this scheme we have focused on the edges aligned with the four main directions along with included the two end pixels in the 5 x 5 window in each direction shown in fig. 1. The impulse detection algorithm is followed.

**Step 1:** The center pixel is classified as noisy by comparing the intensity value of that pixel with the maximum and minimum intensity values of its neighborhood pixels. The method first finds the maximum and minimum intensity values in the 5 x 5 window of the test pixel. If the test pixel does not lie within the intensity range spread of its neighbors it is detected as impulses. Otherwise it is assumed that it may not be impulses and passed to next level detection rule. Let $y_{i,j}$ is the test pixel and $W_{min}$ and $W_{max}$ be the maximum and minimum intensity values respectively within the test window around $y_{i,j}$. Thus the detection of noisy pixel is given as

$$y_{i,j} = \begin{cases} \text{Noisy pixel: } W_{min} \geq y_{i,j} \geq W_{max} \\ \text{Undetected: } W_{min} < y_{i,j} < W_{max} \end{cases} \tag{1}$$

**Step 2:** Let $S_k$ (k=1 to 4) denotes a set of seven pixels aligned in $k^{th}$ direction, origin at (0, 0), i.e,

$S_1$= {(-1,-2), (-2,-2), (-1,-1), (0, 0), (1, 1), (2, 2), (1, 2)}

$S_2$= {(1,-2), (0,-2), (0,-1), (0, 0), (0, 1), (0, 2), (-1, 2)}

$S_3$= {(2,-1), (2,-2), (1,-1), (0, 0), (-1, 1), (-2, 2), (-2, 1)}

$S_4$= {(-2,-1), (-2, 0), (-1, 0), (0, 0), (1, 0), (2, 0), (2, 1)}.

Then let $S_k^0 = S_k \backslash (0, 0)$, $\forall$ k from 1 to 4.

**Step 3:** In 5 x 5 window centered at (i, j), in each direction, define $d_{i,j}^{(k)}$ the sum of all absolute differences of intensity values between $y_{i+s,j+t}$ and $y_{i,j}$ with (s, t)$\in$ $S_k^0$ (k= 1 to 4), given in eq. 2.

**Step 4:** In each direction, weigh the absolute differences between two closest pixels from the center pixel with a large $\omega_m$, weigh the absolute differences between two corner pixels from the center pixel with $\omega_n$ and weigh the absolute differences between two far pixels from the center pixel with a small $\omega_o$, before calculate the sum. Assign $\omega_m$=2, $\omega_n$ = 1 and $\omega_o$= 0.5.

Thus we have, $d_{i,j}^{(k)} = (\sum_{(s,t)\in s_k^0} \omega_{s,t} \mid y_{i+s,j+t} - y_{i,j} \mid, 1 <= k <= 4)$ \hfill (2)

Where $\omega_{s,t} = \begin{cases} \omega_m\text{: } (s, t) \in \Omega^3 \\ \omega_o\text{: } (s, t) \in \Omega^2 \\ \omega_n\text{: otherwise} \end{cases}$ \hfill (3)

Where $\Omega^3$ = {(s, t):-1 $\leq$ s, t$\leq$ 1 }, and \hfill (4)

$\Omega^2$ = {(s, t): (s, t) = $\pm$ {(-1, -2), (1, -2), (2, -1), (-2, -1)}} \hfill (5)

**Step 5:** $d_{i,j}^{(k)}$ is termed as direction index. Find the minimum of these four direction indices, which is used for impulse detection, denoted as

$r_{i,j} = $ min{ $d_{i,j}^{(k)}$ : 1$\leq$ k$\leq$ 4 } \hfill (6)

There may be three cases for value of $r_{i,j}$.

    1. $r_{i,j}$ is small when the center pixel is on a noise free flat region.

2. $r_{i,j}$ is small when the center pixel is on the edge.

3. $r_{i,j}$ is large when the center pixel is a noisy pixel.

**Step 6:** So from the above analysis, classify the center pixel by introducing a threshold T.

Define the complete impulse detection rule as

$y_{i,j}$ is a $\left\{ \begin{array}{l} \text{Noisy pixel: } W_{min} \geq y_{i,j} \geq W_{max} \\ \text{Noise free pixel: } r_{i,j} \leq T \text{ and } W_{min} < y_{i,j} < W_{max} \end{array} \right.$ (7)

## 3. IMPULSE DETECTOR

In the proposed technique a novel scheme has been introduced which is based on minimum variance of all the four directional pixels. The followings are the procedure to restore a noisy pixel $y_{i,j}$ using its neighbourhood pixels.

**Step 1:** Calculate the standard deviation, $\sigma_{i,j}^{(k)}$ of all $y_{i+s,\,j+t}$ with $(s, t) \in S_k^0$, k=1 to 4

**Step 2:** Find the minimum of $\sigma_{i,j}^{(k)}$ : k=1 to 4, as

$$l_{i,j} = \frac{min}{k} \{ \sigma_{i,j}^{(k)}: k=1 \text{ to } 4\} \qquad (8)$$

**Step 3:** Select the set of pixels in the $l_{i,j}$ direction as S. And replace the middle pixel by a variable x to form S = {a, b, c, x, d, e, f}. (9)

**Step 4:** Formulate a quadratic equation f(x) by calculating the variance ( $\sigma^2$ ) of the above set, given in eq. 10. So

$$f(x) = (a - mean)^2 + (b - mean)^2 + (c - mean)^2 + (x - mean)^2$$
$$+ (d - mean)^2 + (e - mean)^2 + (f - mean)^2 \qquad (10)$$

$$where\, mean = (a + b + c + x + d + e + f) / 7 \qquad (11)$$

**Step 5:** Compute first order derivative (f´ (x)) and second order derivative (f´´ (x)) of f(x).

**Step 6:** By the principle of maxima/minima on a quadratic equation and where a, b, c, d, e and f are positive integer constants, the value of f'(x) is always positive for any value of x, where x ∈ [0,255]. So solve the equation f'(x) =0, and get an x, where x ∈ [0,255] for which f(x) is minimum.

**Step 7:** Replace $y_{ij}$ by x.

The methods of detection and filtering of noisy pixels discussed work with three important user parameters. These are number of iterations (I), threshold value (T) and decreasing rate (R) of threshold value in each iteration. These parameters I, T and R are estimated to get optimum restoration results by a population based randomized search technique. Using this technique, the detection and filtering algorithm does not require any parameter to be supplied by the user for any level of noise density in the image.

# 4. PSO BASED OPTIMIZATION

In this paper, a biologically-inspired evolutionary computation (EC) techniques motivated by a social analogy has been incorporated. Particle swarm optimization (PSO) is a population based stochastic optimization technique developed by Dr. Eberhart and Dr. Kennedy in 1995 [18], inspired by social swarming behaviour of bird flocking, fish schooling or even in human social behaviour, from which the swarm intelligence (SI) paradigm has been developed [19]. The main strength of PSO is its fast convergence and easy implementation. The system is initialized with a population of random solutions and searches for optima by updating generations. In PSO, the potential solutions, called particles, fly through the problem space by following the current optimum particles. The search is continued either for fixed number of iterations or till some criteria of optimum solutions based on fitness value is met. Each particle keeps track of its coordinates in the problem space which are associated with the best solution (fitness) it has achieved so far. This value is called *pBest*. Another "best" value that is tracked by the particle swarm optimizer is the best value, obtained so far by any particle in all the population as its topological neighbours, the best value is a global best and is called *gBest*. The particle swarm optimization concept consists of, at each time step, changing the velocity of (accelerating) each particle toward its *pBest* and *gBest* locations. Acceleration is weighted by a random term, with separate random numbers being generated for acceleration toward *pBest* and *gBest* locations. The problem formulation based on PSO model in the supervised way has been resented in next subsection.

## 4.1. Performance Metric

As the maximum value of *PSNR* to be estimated using eq. 12, same equation is used as fitness function f for the particles in PSO based optimization technique.

$$f = PSNR\ (I_1, I_2) = 10 * \log_{10}\left(\frac{255^2}{\frac{1}{M*N}\sum_{m,n}(I_1(m,n) - I_2(m,n))^2}\right) \qquad (12)$$

where $M$ and $N$ are the dimensions of the input images respectively. $I_1$ and $I_2$ are the original and enhanced images respectively.

The detection of noisy pixels depends upon a threshold value $T$, which is decreased by a rate $R$ and the finite numbers of iterations are required to obtain the optimum fitness value depending upon the parameter $R$ and $I$ respectively. The problem under consideration is to find the particles having the best fitness value (i.e., maximum *PSNR*) and that has been implemented in supervised way using the algorithm given in section 4.2.

## 4.2. PSO based optimization algorithm

**Step 1**: Three dimensional search space represented through the attributes $I$, $T$ and $R$ as parameters and initialized 3 to 6, 300 to1000 and 0.6 to 0.95 respectively. Particles are initialized randomly at $x_p$ in a fixed size of swarm. Here '$p$' represents particle number in a swarm. Swarm size is considered here of 6 to 10 particles. At the initial position $x_p$, fitness values $f_p$ are evaluated for individual particle using eq. 12.

**Step 2:** The updated positions $x_p(i+1)$ of the particles are evaluated on calculating the velocities of each particle $v_p(i+1)$ in the search space using eq. 13 and 14.

$v_p(i+1) = h(i)v_p(i) + \Psi_p * r_p * (x_{gbp}(i) - x_{pn}(i)) + \Psi_g * r_g * ((x_{gbp}(i) - x_p(i))$ (13)

$$x_p(i+1) = x_p(i) + v_p(i+1) \hspace{3cm} (14)$$

variables and constants of the above equations are initialized as follows:

1. $\Psi_p$ and $\Psi_g$ are the positive learning factors respectively. Usually $\Psi_p$ equals to $\Psi_g$ and ranges from [0, 4]. Present implementation considered $\Psi_p$ and $\Psi_g > 1$.

2. $r_p$ and $r_g$ random numbers in [0, 1], generated in every iteration separately. They are the real constants used to maintain the diversity of the populations.

3. i is the iteration number initialized to 1 and $I_{MAX}$ is the desired maximum number of generations. In the experimentation, it is set to [10- 20].

4. $h(i)$ are the inertia factors, which has positive real random values in less than 1. This value is kept fixed for individual iteration.

5. $x_p(i)$ and $v_p(i)$ are position and velocity of the $p^{th}$ particle at $i^{th}$ iteration, respectively. Initial positions of particles are randomly initialized and initial velocities are initialized to zero as discussed earlier.

6. $f_{pB}(i)$ and $f_{gB}(i)$ are the *pBest* (personal best fitness value of a particle) value and *gBest* (global best fitness value of particles) values at $i^{th}$ iteration, respectively. Initially $f_{pB}(i)$ are the values of $f_p$ which is calculated in step 1 and the best value among the initialized $f_p(i)$ is the global best initialized value which is assigned to all particles as $f_{pB}(i)$.

7. $x_{pB}(i)$ and $x_{gB}(i)$ are the personal best positions and the global best position of *pth* particle at $i^{th}$ iteration, respectively. These values are initialized by assigning location of particle where $f_{pB}(i)$ and $f_{gB}(i)$ have been obtained respectively, in step 6.

**Step 3:** The velocities and positions of particles are updated using eqns. 13 and 14 respectively. These velocities and positions are calculated using three components; current velocity of each particle, distance between its current position and its *pBest* position of each particle and distance between its current position and *gBest* position of the entire swarm particle.

**Step 4:** To keep the new positions in the search boundary, it is set to [$v_{Min}$, $v_{Max}$]. If new positions of particles are found beyond the boundaries of search space then they are restricted to the boundary values of the search space. The boundary values of I, T and R is discussed in step 1.

**Step 5:** The $f_p(i+1)$ calculated in step 4 is compared with its previous $f_{pB}(i)$. If $f_p(i+1)$ is better than previous $f_{pB}(i)$ then $f_{pB}(i+1)$ is updated by $f_p(i+1)$, otherwise old $f_{pB}(i)$ is retained as a current $f_{pB}(i+1)$. Similarly $x_{pB}(i+1)$ is also updated according to this updated fitness $f_{pB}(i+1)$.

**Step 6:** Best value among the all current $f_{pB}(i+1)$ calculated in step 5 is considered as new $f_{gB}(i+1)$. If new value of $f_{gB}(i+1)$ is better than previous $f_{gB}(i)$ then values of $f_{gB}(i)$ is updated by new $f_{gB}(i+1)$, otherwise old $f_{gB}(i)$ is retained as new $f_{gB}(i+1)$. Similarly, $x_{gB}(i+1)$ is also updated according to this updated fitness $f_{gB}(i+1)$.

**Step 7:** Steps 3 to 6 is repeated until an adequate fitness is reached or a desired maximum number of iterations are met, but for present implementation the interval [10, 20] is taken as steps for iteration.

## 5. SIMULATIONS

The proposed impulse detection, filtering and optimization using particle swarm optimization techniques discussed in previous section is implemented and the performance of the proposed algorithm is simulated on various bench mark images like *Boats, Bridge, Lena* and *Baboon* corrupted by various noise ratios. All test images have the dimensions of 512 x 512 and 8-bit gray levels. The proposed filter is experimented to see how well it can remove the random valued impulses and enhance the image restoration performance for signal processing. These extensive experiments have been conducted to evaluate and compare the performance of the proposed PSO based optimization filter with a number of existing impulse removal techniques. The proposed algorithm have been executed on the machine configuration as ACPI uni-processor with Intel® Pentium® E2180 @ 2.00 Ghz CPU and 2.98 Gbyte RAM with MATLAB 8a environment.

## 5.1. Results and Comparisons

To compare the restoration results of proposed operator with various existing operators each of *Lena*, *Boat* and *Bridge* images corrupted with 40%, 50% and 60% noise densities respectively are taken into account. Using the proposed algorithm on these nine images restoration results are obtained and average *PSNR* values obtained are given in table 1, table 2 and table 3 respectively. It is seen from these tables that the performance of the ACWM [4] is the worst of all in all the cases. The MSM [10] is considerably better than the ACWM [4] in all the cases but worse than the others. The performances of the SD-ROM [13] and PWMAD [8] are very close to each other in all the three types restoration cases. The performances of the DWM [14] operator shows that this filter works better than any existing filter in restoring 40% or more corrupted images. The ANDWP [22] operator also gives excellent restoration results. But the proposed filter obtained very good results (average *PSNR*) for all the images taken in de noising highly corrupted images.

Fig. 2 shows the restoration images in enlarged form to show the preservation of fine details using various filters. For this purpose *Baboon* is taken as test image corrupted by 25% random value impulse noise. It is observed from this figure that the performance of the SMF [1] and MSM [10] are very close to each other. Some noise patches are easily visible in the output images of these two filters. The output images of the SD-ROM [13], FF [16], and PSM [11] are almost indistinguishable from each other and they are significantly better than those of the SMF [1] and MSM [10]. SD-ROM [13], FF [16], and PSM [11] filters show very good noise removal performance but considerably blur the fine details of the image. It is seen that the proposed operator yields much better detail in terms of preservation.

Restoration results in output images by different filters along with the proposed filter on 60% corrupted *Lena* image is given in Fig. 3. We can see from this figure that the output image by MSMF [10] cantains maximum black pathes and performs worst. SD-ROM [13] and PWMAD [8] performs better than MSM [10] but not so well as these have also noise in the reconstructed images respectively. On the contrary DWM filter[14] performs good as it preserves the fine details but can not remove all the patches on the enhanced image. From these restoration results we can see that the proposed operator performs quite well. It has removed almost all the noisy pixels with preservation of image details.

Table 4 shows the performances of the proposed operator in comparison to other filters. The noise densities used here from 20% to 60% with 10% increments. It is seen from this table, the performances of the SMF [1] operator is very poor when the PSM [11] is much better than that in restoring only 20% noise density but for other noise densities it is better but not so good. The ACWM [4], MSM [10], SD-ROM [13] and Iterative median [20] perform very similar way. SD-ROM [13] performs optimally among them in restoring only 50% and 60% noise densities. The PWMAD [8] is better than second order filter [15] in all cases except the 60 % case. The DWM [14] operator performs best than any existing filter in all cases. The ANDWP [22] operator also performs excellently with restoration results. But the proposed filter performs significantly better than any existing filter in restoring 40% or more corrupted images.

Table 1
Average *PSNR* (dB) values for 40% noise density

|  | *Lena* | *Boat* | *Bridge* | *Average* |
|---|---|---|---|---|
| ACWM[4] | 28.79 | 26.17 | 23.23 | 26.06 |
| MSM[10] | 29.26 | 25.56 | 23.55 | 26.12 |
| SD-ROM[13] | 29.85 | 26.45 | 23.8 | 26.7 |
| PWMAD[8] | 31.41 | 26.56 | 23.83 | 27.26 |
| DWM Filter[14] | 32.62 | 27.03 | 24.09 | 27.91 |
| ANDWP[22] | 32.65 | 29.23 | 26.38 | 29.42 |
| **Proposed** | 32.88 | 29.33 | 26.57 | 29.59 |

Table 2
Average *PSNR* (dB) values for 50% noise density

|  | *Lena* | *Boat* | *Bridge* | *Average* |
|---|---|---|---|---|
| ACWM[4] | 25.19 | 23.92 | 21.32 | 23.47 |
| MSM[10] | 26.11 | 24.27 | 22.03 | 24.13 |
| SD-ROM[13] | 26.8 | 24.83 | 22.42 | 24.68 |
| PWMAD[8] | 28.5 | 24.85 | 22.2 | 25.18 |
| DWM Filter[14] | 30.26 | 25.75 | 23.04 | 26.35 |
| ANDWP[22] | 30.50 | 28.72 | 25.51 | 28.24 |
| **Proposed** | 30.91 | 28.92 | 25.62 | 28.48 |

Table 3
Average *PSNR* (dB) values for 60% noise density

|  | *Lena* | *Boat* | *Bridge* | *Average* |
|---|---|---|---|---|
| ACWM[4] | 21.19 | 21.37 | 19.17 | 20.57 |
| MSM[10] | 22.14 | 22.21 | 20.07 | 21.47 |
| SD-ROM[13] | 23.41 | 22.59 | 20.66 | 22.22 |
| PWMAD[8] | 24.3 | 22.32 | 20.83 | 22.48 |
| DWM Filter[14] | 26.74 | 24.01 | 21.56 | 24.10 |
| ANDWP[22] | 28.29 | 26.95 | 23.42 | 26.22 |
| **Proposed** | 28.53 | 26.96 | 23.76 | 26.41 |

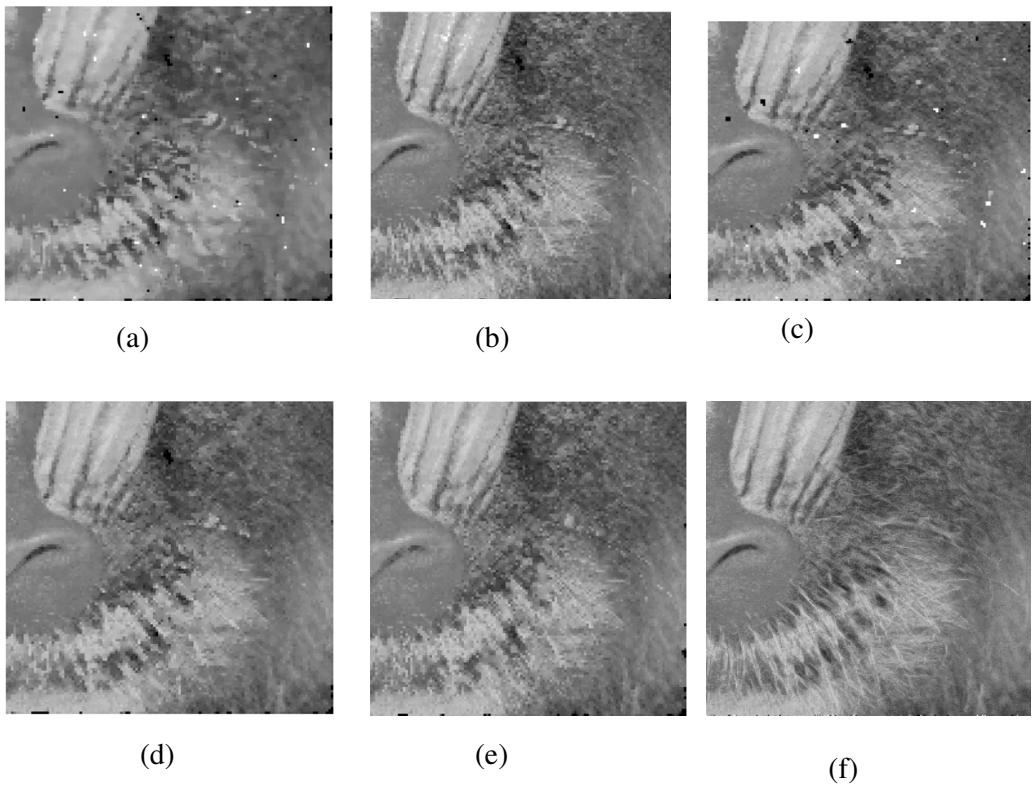

Fig.2 (a) SMF [1] (b) FF [16] (c) MSM [10] (d) SD-ROM [13] (e) PSMF [11] (f) **Proposed**.

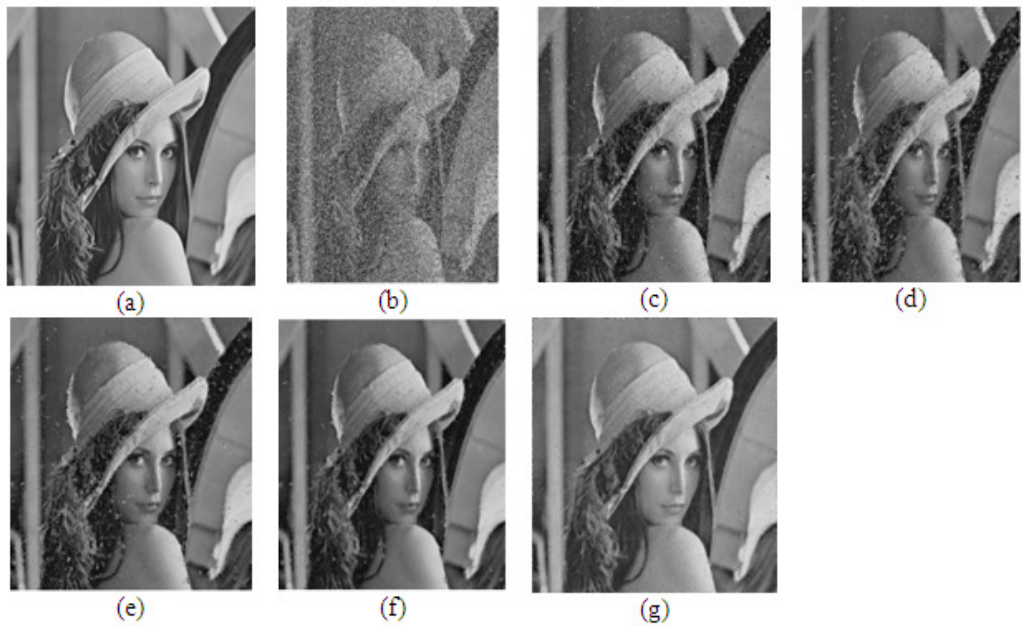

Fig.3 (a) Original (b) 60 % Noisy (c) SD-ROM [13] (d) MSM [10] (e) PWMAD [8] (f) DWM [14] (g) **Proposed**.

Table 4

*PSNR* (dB) values obtained against different noise densities on *Lena* image

| Filter Name | 20% | 30% | 40% | 50% | 60% |
|---|---|---|---|---|---|
| SMF[1] | 30.37 | 30 | 27.64 | 24.28 | 21.58 |
| PSM[11] | 35.09 | 30.85 | 28.92 | 26.12 | 22.06 |
| ACWM[4] | 36.07 | 32.59 | 28.79 | 25.19 | 21.19 |
| MSM[10] | 35.44 | 31.67 | 29.26 | 26.11 | 22.14 |
| SD-ROM[13] | 35.72 | 30.77 | 29.85 | 26.80 | 23.41 |
| Iterative Median [20] | 36.90 | 31.76 | 30.25 | 24.76 | 22.96 |
| Second  Order[15] | 34.35 | 32.53 | 30.90 | 28.22 | 24.84 |
| PWMAD[8] | 36.50 | 33.44 | 31.41 | 28.50 | 24.30 |
| DWM Filter[14] | 37.15 | 34.87 | 32.62 | 30.26 | 26.74 |
| ANDWP[22] | 34.42 | 33.01 | 32.65 | 30.50 | 28.29 |
| **Proposed** | 34.53 | 33.12 | 32.88 | 30.91 | 28.53 |

## 5.2. Comparison of Sensitivity and Specificity

The *miss* and *false* are two measures of performance of noise detection. The *miss* counts the actual numbers of noisy pixels those are not counted. The *false* parameter measures the numbers of noise free pixels which are identified as noisy pixels. A theoretical optimal result can achieve zero *miss* and zero *false* values. Using the proposed PSO based noise removal algorithm, the *miss* and *false* values on 40%, 50% and 60% noisy *Lena* images are given in table 5. We can see from table 5 that SD-ROM [13] and ACWM [4] filter performs excellent for identifying *false* values but it performs very poor for identifying noisy pixels and these undetected noisy pixels become the noticeable patches on the reconstructed image. The ANDWP [22] operator also gives excellent miss and false results.  From table 5 it is also seen that the proposed algorithm can identify the noisy pixels as well as it can free pixels correctly with a remarkable difference compared to all other existing filters. It gives optimum *miss* and *false* values among all filters taken into account for the experiment.

Two other statistical measurement tools of performance are also used to measure the performance of proposed algorithm. These are *sensitivity (Sen#)* and *specificity (Spc#)*. *Sensitivity* measures the proportion of positives which are correctly identified as such. *Specificity* measures the proportion of negatives which are correctly identified. 100% *sensitivity* and 100% *specificity* are the optimal results.

It is seen from table 6 that the sensitivity and specificity for different conventional filters along with the proposed for 40%, 50% and 60% corrupted *Lena* images, proposed algorithm obtain very good results in terms of sensitivity and specificity.

Table 5

Comparison of *miss* and *false* results for "*Lena*" image

| Filter | 40% | | 50% | | 60% | |
|--------|------|-------|------|-------|------|-------|
| | **miss** | **false** | **miss** | **false** | **miss** | **False** |
| SD-ROM[13] | 22842 | 411 | 32566 | 998 | 45365 | 2651 |
| MSM[10] | 16582 | 7258 | 20857 | 10288 | 26169 | 15778 |
| ACWM[4] | 16052 | 1759 | 23683 | 2895 | 32712 | 7644 |
| PWMAD[8] | 11817 | 9928 | 14490 | 15003 | 17760 | 19577 |
| DWM[14] | 9512 | 7761 | 9514 | 11373 | 12676 | 12351 |
| ANDWP[22] | 7852 | 6018 | 8260 | 7512 | 8812 | 9304 |
| **Proposed** | 7602 | 5836 | 8066 | 7452 | 8565 | 9158 |

Table 6

Comparison of *sensitivity* and *specificity* results for "*Lena*" image for different noise densities

| Filter | 40% | | 50% | | 60% | |
|--------|----------|----------|---------|---------|---------|----------|
| | **Sen# %** | **Spc# %** | **Sen# %** | **Spc# %** | **Sen# %** | **Spc# %** |
| SDROM[13] | 78 | 99 | 72 | 99 | 71 | 98 |
| MSM[10] | 84 | 97 | 84 | 92 | 83 | 89 |
| ACWM[4] | 84 | 98 | 81 | 97 | 79 | 95 |
| PWMAD[8] | 88 | 90 | 88 | 88 | 88 | 87 |
| DWM[14] | 90 | 92 | 92 | 91 | 91 | 92 |
| ANDWP[22] | 93 | 94 | 94 | 94 | 94 | 94 |
| **Proposed** | 93 | 93 | 93 | 93 | 94 | 94 |

## 6. CONCLUSIONS

In this paper, a novel approach has been presented for filtering high random valued impulse noise from digital images. In this approach tuning parameters of noise detection and filtering operator has been optimized in supervised way using PSO based optimization technique. The main advantage of the proposed operator over most other operators is that it efficiently removes impulse noise from digital images while successfully preserving thin lines, edges and fine details in the enhanced image.

**Authors:**

**Jyotsna Kumar Mandal,** M. Tech.(Computer Science, University of Calcutta),Ph.D.(Engg., Jadavpur University) in the field of Data Compression and Error Correction Techniques, Professor in Computer Science and Engineering, University of Kalyani, India. Life Member of Computer Society of India since 1992 and life member of cryptology Research Society of India. Dean Faculty of Engineering, Technology & Management, working in the field of Network Security, Steganography, Remote Sensing  & GIS Application, Image Processing. 25 years of teaching and research experiences. Eight Scholars awarded Ph.D. and 8 are pursuing. Total number of publications 189.

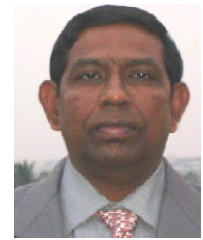

**Somnath Mukhopadhyay** did his graduation in Bachelor of Computer Application form the University of Burdwan in 2004 and MCA in 2008 from the same university. He did his M.Tech (CSE) in 2011 from the University of Kalyani. Currently he is engaged in teaching profession with three years of experience. Broad area of his research interest includes Signal and Image Processing, Bioinformatics and Pattern Recognition. He has 7 publications in international conference proceedings and journals.

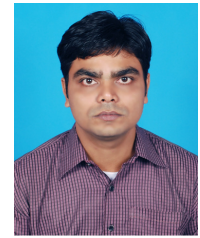